\documentclass[11pt]{article}

\usepackage{graphicx}
\usepackage{enumitem}
\usepackage{placeins}
\usepackage{natbib}
\usepackage{booktabs}
\usepackage{amsmath,amssymb}
\usepackage{graphicx}

\usepackage[letterpaper,margin=1in]{geometry}
\usepackage{longtable}
\usepackage{array}
\usepackage{tabularx}
\usepackage{microtype}
\usepackage[hidelinks]{hyperref}
\usepackage{xurl}

\newcolumntype{Y}{>{\raggedright\arraybackslash}X}

\setlist[itemize,1]{leftmargin=1.35em,label=\textbullet}
\setlist[itemize,2]{leftmargin=1.35em,label=--}
\setlist[itemize,3]{leftmargin=1.35em,label=*}
\setlist[itemize]{itemsep=0.35em,topsep=0.35em,parsep=0pt,partopsep=0pt}

\newcommand{\model}[1]{\texttt{#1}}

\title{The Judge Is Not Impartial: Self-Preference in Medical LLM Evaluation}

\author{%
\parbox{0.82\textwidth}{%
\centering
\textbf{Zara Ansari}\textsuperscript{*1,2},
\textbf{Shlok Natarajan}\textsuperscript{*1},
\textbf{Vincent Yip}\textsuperscript{3},
\textbf{Cally Lin}\textsuperscript{1},
\textbf{Avi Udash}\textsuperscript{3},
\textbf{Mia Garvey}\textsuperscript{3},
\textbf{Aaron Fanous}\textsuperscript{1}, and
\textbf{Roxana Daneshjou}\textsuperscript{1,4}
\\[0.8em]
{\small
\textsuperscript{1}Department of Biomedical Data Science,
Stanford University, Stanford, CA 94305, USA.
\\
\textsuperscript{2}Department of Computer Science,
Harvey Mudd College, Claremont, CA 91711, USA.
\\
\textsuperscript{3}Department of Computer Science,
Stanford University, Stanford, CA 94305, USA.
\\
\textsuperscript{4}Department of Dermatology,
Stanford University, Stanford, CA 94305, USA.
\\[0.5em]
}%
}%
}%

\date{September 2026}

\begin{document}

\maketitle
\footnotetext[1]{Indicates equal contribution}

\begin{abstract}

Large language models (LLMs) are increasingly used to evaluate model-generated medical content, creating a potential circularity: an LLM judge may favor outputs generated by the same model independently of clinical quality. To characterize LLM judge self-preference in medical settings, we measured same-model affinity in single-turn specialist question answering and multi-turn standardized-patient encounters. We evaluated eight models, two from OpenAI, Anthropic, Google, and Qwen, on 620 real point of care questions from clinicians (Real-POCQi). The same eight models also generated and judged 200 standardized patient-clinician scenarios (MedSP1000) up to 8 visible turns. In total, our analysis spans 13,760 LLM judgments. 

We define three complementary measures of self-preference. Answer Rank Bias, our primary measure, compared the rank assigned to a fixed response by its own-model judge with its average rank from outside judges. Answer Score Bias made the analogous comparison using rubric scores. Own-Answer Win Rate measured how often a judge ranked its own model's response above a competitor. Higher values across all these measures indicate greater degrees of self-preference. In Real-POCQi, pooled Answer Rank Bias was $+1.22$ positions, and all eight judges favored their own answers. Self-preference persisted both with and without a grading rubric. Revealing generator identities modestly weakened self-preference by an average of $-0.094$ positions. In MedSP1000, pooled Answer Rank Bias ranged from $+0.53$ to $+0.62$ positions, with no clear difference by conversation length ($p=0.17$).

LLM judges systematically favored responses generated by their own models across multiple formats. Clinical AI deployments should exercise increased vigilance including multiple evaluation measures and judge panels spanning differing model families when deploying LLM-as-a-judge systems.

\end{abstract}
\pagebreak
\section{Introduction}

As large language models (LLMs) are considered for different tasks in medicine, benchmarking has become a key method for choosing the best model for a particular task.  However, the complexity of medical practice has called for larger and more complex benchmarking to compare the performance of AI models. Human review of thousands of scenarios tested across multiple AI models is untenable due to cost and time \citep{zheng2023judging}. A recent review of medical AI benchmarking found that large language models (LLMs) are increasingly used as judges to evaluate the outputs of other LLMs (and in some cases, their own outputs) \citep{li2026llm}. This same study found that OpenAI's models are currently the most popular judges for medical AI benchmarking studies \citep{li2026llm}.

Researchers and industry experts claim that using an LLM as a judge provides a scalable way to evaluate responses based on criteria such as accuracy, clinical utility, source quality, verifiability, and completeness.  The stakes in medical AI evaluation are high since biased judges could incorrectly favor unsafe models and influence clinical deployment decisions to the detriment of patients. Furthermore, evaluating medical responses is especially complicated since an answer may sound confidently correct, but still contain incomplete medical information. As a result, evaluating LLMs for medical tasks often requires several dimensions of response quality to be considered simultaneously.

Prior work in general generative AI tasks has shown that LLM judges may exhibit biases such as positional and verbosity bias. \cite{wataoka2025selfpreference} examined how LLM judges can favor responses that reflect policies or writing styles familiar to them.  In fact, one study showed that LLM judges may favor AI generated clinical responses over human produced ones. \cite{alvarezarenas2026unreliable} found that this preference seemed to depend more on how an answer was written, especially how long it was and the language used, than on whether the model could actually tell if the answer came from a human or an AI. In one experiment, long responses still received high scores even when matched with unrelated questions, demonstrating that verbosity can outweigh content for some LLM judges. \cite{wataoka2025selfpreference} introduced a quantitative way to measure self-preference in LLM judges. They found that GPT-4 showed significant self-preference and that LLM judges tended to rate lower perplexity responses more highly than human evaluators did.  Work from \cite{pombal2026selfpreference} looked at whether rubric-based evaluations (with a set of yes or no criteria) could mitigate the self-preference of LLM judges. Their work found that LLM self-preference persists even with fully "objective" rubrics.

Self-preference in medical contexts is understudied. To address these gaps, we evaluate self-preference across four model families in single-turn generalist-specialist question answering and in the multi-turn patient–physician conversations. In order to address the fact that a model may choose its own response because its answer is genuinely better, we evaluate the models preference in three distinct ways. Our primary measure is Answer Rank Bias, which compares how a model's own judge and outside judges rank the same fixed response. We complement this measure with Answer Score Bias, which quantifies how different judges score the same fixed response, and Own-Answer Win Rate, which measures how often a judge ranks its own model's response above a competing response. Our findings of LLM judge self-preference in medical contexts provide important considerations for medical AI benchmarking and monitoring.

\section{Methods}


Real-POCQi contains 620 real world clinical questions which were submitted by practicing U.S. physicians through the OpenEvidence platform. The questions span 30 medical specialties and focus on point-of-care clinical decision making. Each entry includes a unique question ID, the question text, and its respective specialty. The dataset was reviewed to remove identifying patient information and other content that could raise privacy concerns. We use all 620 questions as prompts for generating and evaluating responses from the eight tested models.

MedSP1000 contains standardized-patient scenarios derived from published clinical teaching cases. Each scenario includes separate materials for the physician and the standardized patient. The dataset also includes evaluator and environment-controller materials, which we exclude because our study focuses on generating and evaluating the consultations. We screened the scenarios for standard patient–physician format, sufficient patient information, English-language content, and the absence of diagnosis-key leakage and caregiver-proxy framing. Of 217 eligible scenarios, we randomly sampled 200 scenarios for our analysis. In this study, we provide the physician-facing information to the simulated clinician model and the patient information to guide the simulated patient model, generating multi-turn patient–physician conversations across the eight tested models. The patient model was held as Mistral-Small-3.1-24B-Instruct-2503 across all runs to avoid same-family pairing with any evaluated model.

\subsection{Sample Size Analysis}

We conducted a power analysis using a two-sided significance level of 0.05 and 90\% power to determine the needed sample sizes for detecting self-preference effects. For the Answer Score Bias analysis, described in the following sections, approximately 119 paired observations were needed to detect a small-to-medium effect ($d=0.30$). For the Own-Answer Win Rate analysis, approximately 114 observations were needed to detect a win rate of 65\%, compared with the 50\% rate expected under no preference. Based on these estimates, the 620 questions used in the single-turn setting and the 200 scenarios used in the multi-turn setting provided more than enough observations for both analyses.

In the single-turn setting, each of the eight models answered all 620 questions, resulting in 4,960 specialist responses (620 questions $\times$ 8 models). These responses were then ranked via LLM-as-a-judge. All eight models also served as judges, with each judge scoring and ranking the full set of eight candidate responses for every question. This resulted in 4,960 rankings (8 judges $\times$ 620 questions). In each judgment, the judge scored all eight responses and then ranked them, with the ranking not required to follow the rubric scores.

In the multi-turn setting, each of the eight models generated an eight-turn conversation for all 200 scenarios, producing 1,600 conversations (200 scenarios $\times$ 8 models). The shorter 2, 4, and 6 turn versions were created by truncating these eight-turn conversations. Model ranking by judges was then performed separately at each of the four conversation lengths using the same eight judges, resulting in 6,400 judgments (8 judges $\times$ 4 lengths $\times$ 200 scenarios). In total, our analysis includes 11,360 identity-blind, order-randomized ranking judgments across the two settings. The no-rubric and identity-revealed checks add 800 and 1,600 judgments, bringing the study total to 13,760.

\subsection{Generator and Judge Models}
The study uses two models from each of four families, shown in
Table~\ref{tab:cohort}. Generator models are fed the original query from either the Real-POCQi or MedSP1000 dataset and produce a response. In the single turn and multi turn pipelines, all eight models act as generators and judges.

These tier labels describe their role in this study and
are not intended as a universal performance ranking.

\begin{table}[htbp]
\centering
\caption{Generator and judge cohort.}
\label{tab:cohort}
\begin{tabular}{lll}
\toprule
Family & Higher-Capability Model & Workhorse/Mid-Tier Model \\
\midrule
OpenAI & \model{gpt-5.6-sol} & \model{gpt-5.6-terra} \\
Anthropic & \model{claude-opus-5} & \model{claude-sonnet-5} \\
Google & \model{gemini-3.1-pro-preview} & \model{gemini-3.7-flash} \\
Qwen & \model{Qwen/Qwen3.5-122B-A10B-FP8} & \model{Qwen/Qwen3.8-27B-FP8} \\
\bottomrule
\end{tabular}
\end{table}
\FloatBarrier

\subsection{Single-turn Pipeline}

This pipeline evaluates specialist model behavior on single-turn questions using all 620 questions from the Real-POCQi dataset. Each question is associated with a clinical specialty, such as cardiology or neurology. For each question, eight models (GPT-5.6 Sol, GPT-5.6 Terra, Claude Opus 5, Claude Sonnet 5, Gemini 3.1 Pro, Gemini 3.7 Flash, Qwen3.5-122B, and Qwen3.8-27B) each independently generate a single answer in the role of the relevant specialist, prompted with that specialty's label. The eight models represent two capability tiers from four model families. Each model generates one response per question, resulting in 4,960 total answers.

These responses are ranked and scored across five criteria: accuracy, clinical utility, source quality, verifiability, and completeness (0–5; overall = their mean). These criteria are adapted from the original Real-POCQi publication \cite{feng2026expert}. For each question, a judge receives all eight responses at the same time. The responses are identity-blind, and their presentation order is randomized independently for each judge and question under a fixed seed. In the same evaluation, the judge also ranks all eight responses from best to worst using its overall clinical judgment. The ranking is not required to follow the rubric scores, and rankings are strict with no ties. All eight models also serve as judges, resulting in 4,960 judgments. We refer to this as the primary condition, in which all 620 questions are judged with both rubric scores and a full ranking, and generator identities are hidden. These judgments yield three measures of self-preference. Answer Rank Bias is our primary measure, and we also report Answer Score Bias and Own-Answer Win Rate (these measures are defined below).

We also evaluated two sensitivity conditions, reusing the same generated responses so that only the judging setup changed. The no-rubric condition tests whether providing the five criteria changes the rankings. All eight judges rank the same blinded responses on overall quality alone,
with no criteria provided, for a random subset of 100 questions. The identity-revealed condition labels every candidate with its generator model for a subset of 200 questions. All eight generator models also serve as judges, and all eight generators remain in each candidate list.

\subsection{Multi-turn Pipeline}

This pipeline evaluates physician-model behavior in multi-turn patient–physician conversations, using 200 scenarios from MedSP1000. Each scenario provides case material written for the examinee, which we give to the physician model. We give a separate private set of instructions written for the standardized patient to the patient model. Format and level of detail vary between cases. The physician model only sees its own material before beginning the consultation followed by the patient replies.

Eight physician models, GPT-5.6 Sol, GPT-5.6 Terra, Claude Opus 5, Claude Sonnet 5, Gemini 3.1 Pro, Gemini 3.7 Flash, Qwen3.5-122B, and Qwen3.8-27B, each independently generate an eight turn conversation for every scenario. Physician turns are produced by the model under study, while patient turns are generated by a standardized patient simulator, Mistral Small 3.1 24B Instruct, which is not evaluated as a generator or judge. Each physician model is given four speaking turns, and the patient model is given four turns, resulting in an eight-turn physician–patient conversation. Each model generates exactly one conversation per scenario, producing 1,600 total conversations. Shorter conversation lengths of two, four, and six turns are derived by truncating each eight turn conversation to the corresponding length. Since each turn depends only on the conversation history that comes before it, truncating the longer conversations provides valid shorter versions.

Similar to the single-turn pipeline, conversations in the multi-turn pipeline are scored and ranked across the same five criteria (0–5; overall = mean) for measurement consistency. These criteria are applied to the physician’s behavior across the encounter. For each scenario, a judge receives all eight conversations at the same time and scores only the physician responses, while the patient turns are included as context but are not scored. The conversations are identity-blind, and their presentation order is randomized independently for each judge and scenario under a fixed seed. In the same evaluation, the judge also ranks all eight conversations from best to worst using its overall clinical judgment. The ranking is not required to follow the rubric scores, and rankings are strict with no ties. All eight physician models also serve as judges to assess self-preference.

The difference from the single-turn pipeline is that the scoring is performed separately at each of the four conversation lengths (2, 4, 6, and 8 turns). This results in 1,600 judgments at each conversation length and 6,400 judgments in total. Thus, the rankings and the three resulting measures of self-preference are computed and reported separately for each model and conversation length.

\begin{figure}[htbp]
    \centering
    \includegraphics[width=0.9\textwidth]{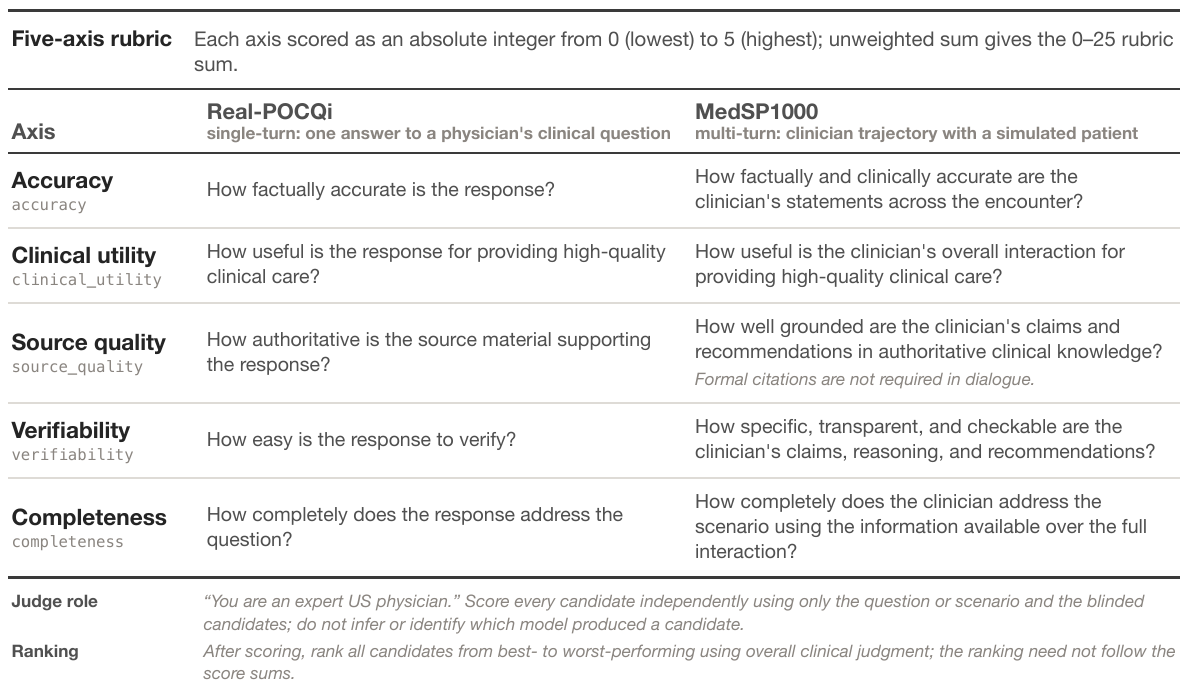}
    \caption{Single and Multi-Turn Scoring Rubrics}
    \label{fig:grading-rubric}
\end{figure}

\section{Analysis of Data}

We summarize model self-preference using three complementary measures: \textit{(1) Answer Rank Bias (ARB)}, our primary measure, holds the exact response fixed and compares the average rank assigned
by outside judges with the rank assigned by the answer’s own-model judge. Positive values indicate greater self-preference (models ranking their own responses in lower numbered positions relative to others). \textit{(2) Answer Score Bias (ASB)} makes the analogous fixed-response comparison using mean rubric scores. Positive values indicate that the judge scores its response more generously. \textit{(3) Own-Answer Win Rate (OAWR)} is the proportion of implied head-to-head comparisons in which a judge ranks its own model's response above a competitor. Unlike the first two measures, the Own-Answer Win Rate compares different responses and, therefore, combines response quality with judging behavior. Larger values of all three measures consistently point toward greater model self-preference.

\subsection{Answer Rank Bias}

Every judge ranks the full candidate list, so each answer receives $J$ ranks: one from the judge belonging to the model that wrote it, and $J-1$ from outside judges. For question $q$ and exact model $j$, let $r_{kqj}$ be the rank that judge $k$ assigns to the answer generated by model $j$, where lower ranks are better. Answer Rank Bias is
\begin{equation}
  D_{jq}
  =
  \frac{1}{J-1}\sum_{k\ne j} r_{kqj}
  -
  r_{jqj}.
  \end{equation}
The first term is the average rank the outside judges give the answer, and the second is the rank the model's own judge gives that same answer. Thus, $D_{jq}>0$ means that judge $j$ ranks its own answer more favorably than outside judges rank the same answer to the same question. Because every judge is ranking the same answer to the same question, the differences in answer quality cancel out. This means that a model that writes strong answers is not credited with self-preference. Conversely, a model whose answers rank poorly can still show self-preference.

Judges can be influenced by the order in which answers are presented, that is, whether an answer appears first, last, or somewhere in between \citep{zheng2023judging}. We measure how much each position shifts a rank, relative to how that judge usually ranks that generator. We then subtract that position effect from every rank before computing the difference. Each judge's value is computed partly from the other judges' ranks, so the $J$ values for a question are related rather than independent. We average them within each question and compute confidence intervals across questions. As a check we repeat the comparison on standardized scores, and where judges are tested separately $p$-values carry Holm correction.

The four MedSP1000 conditions show the same $1{,}600$ conversations at increasing length, so every conversation contributes a value at all four lengths. Length comparisons are thus paired within question. If we write $D^{(2)}$ through $D^{(8)}$ for the value of $D_{jq}$ at each length, then $D^{(4)}-D^{(2)}$ (4 minus 2 turns) is the change in self-preference for a question when four turns are visible relative to two. 
We first ask whether self-preference changes at all across the four lengths. Answering this with three separate tests (4-minus-2, 6-minus-2, and 8-minus-2), one for each change relative to two turns, would inflate the chance of a false positive. Instead we test the three changes together using Hotelling's $T^{2}$, a multivariate paired $t$-test. The null hypothesis, $H_0$, is that all three changes are zero on average, which would mean self-preference is the same at every length.


\subsection{Answer Score Bias}
Let $s_{kqj}$ represent the mean of judge $k$’s five rubric scores for the answer generated by model $j$ on question $q$ (which uses the same $0$--$5$ scale as the criteria). To measure Answer Score Bias, we hold the answer fixed and compare the score assigned by the model’s own judge with the average score assigned by the outside judges:

\begin{equation}
B_{jq}=s_{jqj}-\frac{1}{J-1}\sum_{k\ne j}s_{kqj}.
\end{equation}

Because all judges evaluate the same response, differences in the quality of the answer itself are held constant. The remaining difference reflects how the judges score that response. A positive raw ASB means the own-model judge scores its response more generously than outside judges; a negative value means it scores the response more harshly.

\subsection{Own-Answer Win Rate}

The rankings contain no ties, so each ranking can also be treated as a set of head-to-head comparisons between the judge's own answer and every competing answer. Let $Q$ represent the number of questions in the condition. The Own-Answer Win Rate for model $j$ is
\begin{equation}
P_j=\frac{\sum_q\sum_{m\ne j}\mathbb{I}(r_{jqj}<r_{jqm})}{Q(J-1)},
\end{equation}
where the numerator counts the comparisons in which judge $j$ ranked its own answer above a competitor and the denominator is the total number of such comparisons, $Q(J-1)$. Unlike Answer Score Bias, this measure does not hold answer quality constant because the competing responses are different. So, a model may rank its own answer higher simply because that answer is actually better. The ranking is produced in the same evaluation as the rubric scores, but it is not required to follow those scores for ranking.

\section{Results}

\subsection{Single-Turn Results}
The main finding is that every judge favored its own model's answers, including judges whose own answers ranked among the weakest in the panel. On average, a judge placed its own answer 1.22 positions closer to first than the other judges placed that same answer (95\% CI $[1.18, 1.26]$). The bias was not explained by answer quality, presentation position, medical specialty, or explicit knowledge of model identity. The effect varied substantially by judge. These findings indicate that self-preference is a result of model based evaluation, and its magnitude depends on the judge and on how self-preference is measured.

\begin{figure}[htbp]
    \centering
    \includegraphics[width=0.8\textwidth]{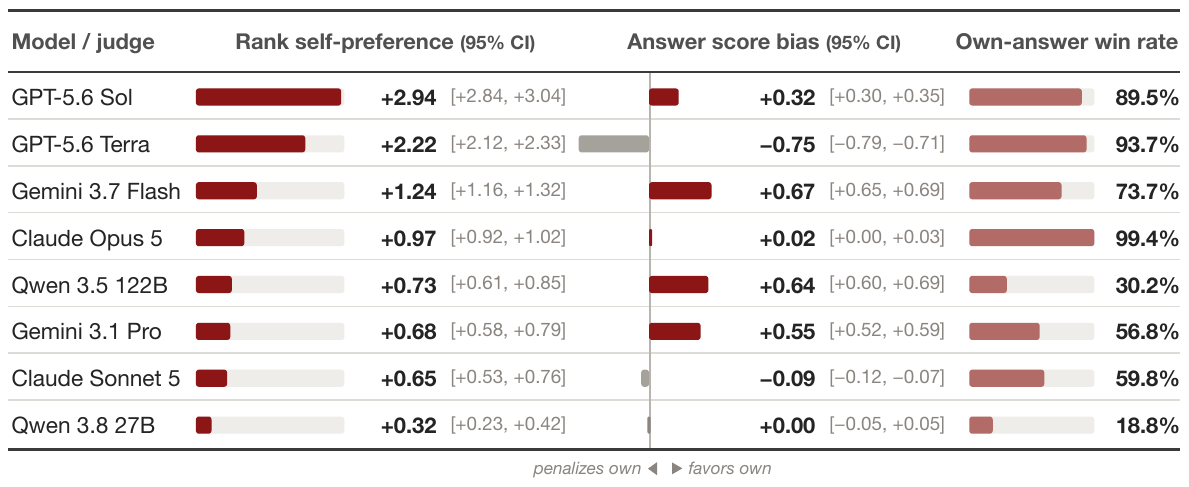}
    \caption{Single-Turn Self-Preference Results}
    \label{fig:single-turn-results}
\end{figure}

\subsubsection{Real-POCQi Generation}

The judges showed a separation in overall generator performance. Claude Opus 5 ranked first overall by a wide margin, with a mean rank of 1.89, and was ranked first in 69.3\% of judgments. The two Qwen models ranked last, with mean ranks of 6.52 and 6.99 and first-place rates below 1\%. The five models between these extremes had a narrow range of mean ranks (3.37 to 4.62), while there was a further 1.90 rank position separation from sixth to seventh place. Thus, the results show a clear divide between the upper six models and the two Qwen models, with relatively small differences among the models in the upper group.

A model's mean rank does not always match how often it comes first. Gemini 3.7 Flash had a better mean rank than GPT-5.6 Sol but was ranked first far less often (1.3\% versus 8.3\%). This suggests that Flash performed consistently well without frequently being the top-ranked model, while Sol’s performance was more variable. Rubric scores make the separation between generators appear smaller. Scores ranged from 16.41 to 23.11 out of 25, so even the lowest-ranked generator received about 66\% of the available points while placing near the bottom of most rankings.


Table~\ref{tab:real-performance} shows how the eight judges collectively ranked the generators. With eight candidates the average rank is 4.5, so values better than 4.5 indicate above-average performance in the judges' collective rankings.

\begin{table}[htbp]
\centering
\small
\caption{Real-POCQi aggregate generator performance across eight
judges. Lower mean rank is better.}
\label{tab:real-performance}
\begin{tabular}{lrrr}
\toprule

Generator & Mean rank & Ranked first & Mean rubric sum \\
\midrule
Claude Opus 5 & 1.89 & 69.3\% & 23.11 \\
GPT-5.6 Terra & 3.37 & 17.5\% & 21.65 \\
Gemini 3.7 Flash & 3.92 & 1.3\% & 21.08 \\
GPT-5.6 Sol & 4.30 & 8.3\% & 20.84 \\
Claude Sonnet 5 & 4.39 & 1.5\% & 20.63 \\
Gemini 3.1 Pro & 4.62 & 1.3\% & 20.41 \\
Qwen 3.5 122B & 6.52 & 0.4\% & 17.52 \\
Qwen 3.8 27B & 6.99 & 0.3\% & 16.41 \\

\bottomrule
\end{tabular}
\end{table}
\FloatBarrier


\subsubsection{Real-POCQi Evaluation}

Every judge ranked its own answer more favorably than the rest of the panel ranked that same answer. On average a judge placed its own answer \textbf{1.22 positions} closer to first (95\% CI $[1.18, 1.26]$), and all eight judge-specific effects remain significant after Holm correction. Correcting for the position in which a candidate was presented leaves this number unchanged, so presentation order does not explain it. In a list of eight an answer can move at most seven positions, so 1.22 is roughly 17\% of the full range; for the strongest judge, GPT-5.6 Sol, it is nearly three positions.

\begin{table}[htbp]
\centering
\small
\caption{Position-adjusted Answer Rank Bias in Real-
POCQi. Positive
values indicate that the judge ranked its own answer more
favorably.}
\label{tab:real-self}
\begin{tabular}{lrr}
\toprule
Judge & Answer Rank Bias & 95\% CI \\
\midrule
GPT-5.6 Sol & +2.94 & $[+2.84,+3.04]$ \\
GPT-5.6 Terra & +2.22 & $[+2.12,+2.33]$ \\
Gemini 3.7 Flash & +1.24 & $[+1.16,+1.32]$ \\
Claude Opus 5 & +0.97 & $[+0.92,+1.02]$ \\
Qwen 3.5 122B & +0.73 & $[+0.61,+0.85]$ \\
Gemini 3.1 Pro & +0.68 & $[+0.58,+0.79]$ \\
Claude Sonnet 5 & +0.65 & $[+0.53,+0.76]$ \\
Qwen 3.8 27B & +0.32 & $[+0.23,+0.42]$ \\
\midrule
Pooled Question Average & +1.22 &
$[+1.18,+1.26]$ \\
\bottomrule
\end{tabular}
\end{table}
\FloatBarrier

The self-preference effect appears across all judges but varies in magnitude. Judge-specific estimates range from $+0.32$ to $+2.94$, and this ordering does not follow generator quality. The two GPT judges show the largest self-preference effects, even though Terra ranks second and Sol fourth in overall performance (Table~\ref{tab:real-performance}). Opus has the strongest performance overall across all judges (Table~\ref{tab:real-performance}), but shows an effect of only 0.97, the fourth largest of the eight. This suggests that self-preference is a property of the individual judge rather than a result of answer quality. The Qwen judges show this clearly. Their answers are the weakest in the panel (Table~\ref{tab:real-performance}), so they almost never win. Their own-answer win rates are also low at 30.2\% and 18.8\% (Table~\ref{tab:real-draft-scores}). Therefore, a measure based on wins alone would call them unbiased. Yet each Qwen model still ranks its own answer above where the outside judges place it (Table~\ref{tab:real-self}).

\begin{table}[htbp]
\centering
\small
\caption{Real-POCQi measures, sorted by Answer Score Bias from highest
to lowest. Answer Score Bias is on the 0--5 mean-rubric scale.
Own-Answer Win Rates convert strict rankings to all
own-answer-versus-competitor decisions.}
\label{tab:real-draft-scores}
\begin{tabular}{lrr}
\toprule
Model/judge & Answer Score Bias (95\% CI) & Own-Answer Win Rate \\
\midrule
Gemini 3.7 Flash & $+0.67;[+0.65,+0.69]$ & 73.7\% \\
Qwen 3.5 122B & $+0.64;[+0.60,+0.69]$ & 30.2\% \\
Gemini 3.1 Pro & $+0.55;[+0.52,+0.59]$ & 56.8\% \\
GPT-5.6 Sol & $+0.32;[+0.30,+0.35]$ & 89.5\% \\
Claude Opus 5 & $+0.02;[+0.00,+0.03]$ & 99.4\% \\
Qwen 3.8 27B & $+0.00;[-0.05,+0.05]$ & 18.8\% \\
Claude Sonnet 5 & $-0.09;[-0.12,-0.07]$ & 59.8\% \\
GPT-5.6 Terra & $-0.75;[-0.79,-0.71]$ & 93.7\% \\
\bottomrule
\end{tabular}
\end{table}
\FloatBarrier

Answer Score Bias and Own-Answer Win Rate diverge from each other and from Answer Rank Bias (Table~\ref{tab:real-draft-scores}), which shows how self-preference is measured matters. Terra has the most negative Answer Score Bias of any judge ($-0.75$), yet it ranks its own answer above 93.7\% of competitors. Opus shows almost no score inflation ($+0.02$), yet it places its own answer above 99.4\% of competitors, which is largely consistent with the fact that outside judges also rank Opus first. Qwen 3.5 122B inflates its own scores more than any judge except Gemini Flash but reaches only a 30.2\% Own-Answer Win Rate, because its answers are far enough behind the other candidates that scoring them generously still leaves them near the bottom of the ranking. These measures disagree because they are not measuring the same thing. For instance, Own-Answer Win Rate depends only on where a judge placed its own answer in its own ranking, and never on how other judges viewed that answer. Answer Score Bias compares the same answer across judges, but it does so in raw rubric points, and a generous judge's half-point is not equivalent to a strict judge's. Answer Rank Bias compares the same answer across judges using ranks, which are defined relative to the same candidate list and are therefore unaffected by how a judge uses the scale (this is why it carries the primary analysis).

\subsubsection{Family-level Preference in Real-POCQi}

Judges from the same family ranked their family's two answers 1.27 positions closer to first than judges from other families ranked those same answers (95\% CI $[1.23,1.31]$). This was similar to the exact-model result of 1.22 positions. The family-level pattern therefore did not disappear when we expanded the comparison from one model to both models in its family.

One possible explanation is that models from the same family share writing styles, reasoning patterns, or ideas about what makes an answer good. However, this analysis cannot separate that possibility from exact-model preference, because the same-family average includes the exact model that generated each answer. Distinguishing the two would require more models per family and repeated answers from each model.

\subsubsection{Ranking Only (No-Rubric) Sensitivity}

Because the primary condition supplies five explicit rubric criteria, one possibility is that judges favor answers written in the style those criteria reward. To test this, the direct-ranking condition asks the same eight judges to rank the same blinded responses on overall quality alone, with no criteria provided. It uses a random sample of 100 questions from the Real-POCQi dataset.

Removing the rubric leads to a pooled effect of $+1.26$ positions (95\% CI $[1.16, 1.36]$), compared to the $1.22$ effect observed when the rubric is used. Seven of the eight judges still favor their own answers by an amount that excludes zero. The only exception is Qwen 3.8 27B, with an effect of $+0.16$ positions (95\% CI $[-0.04, +0.35]$). It also had the smallest effect in the main condition, and this estimate is based on 100 questions rather than 620. The two conditions still produce very similar rankings overall. For any pair of candidates, the rubric and no-rubric conditions agree on which response is better 91.5\% of the time, and they select the same top answer 87.1\% of the time. However, they produce the exact same full eight-answer ranking only 16.1\% of the time. In other words, removing the rubric does not change which answers judges treat as strong or weak; it only reshuffles answers that are already close in quality.

\subsubsection{Identity-Revealed Comparison}

The primary conditions hide generator identity, which means that any self-preference detected there came from the judge ranking only the outputs without knowing which model generated which response. The identity-revealed condition tests the opposite setup by naming the generator of every candidate. This condition covers 200 randomly sampled questions from the Real-POCQi dataset, judged by all eight judges, giving 1,600 judgments, where all eight candidates are still shown, each labeled with its generator. The blinded comparison uses exactly the same 200 questions and the same eight judges, so the only difference between the two conditions is whether names were visible. Presentation-position effects are estimated separately in each condition.

\begin{table}[htbp]
\centering
\small
\caption{Eight-judge position-adjusted identity comparison, sorted by
the change in Answer Rank Bias from the blinded to the identity-revealed
condition. Negative changes indicate weaker self-preference when
generator names are visible.}
\label{tab:identity}
\begin{tabular}{lrrr}
\toprule
Judge & Revealed ARB & Blinded ARB & Change (revealed $-$ blinded) \\
\midrule
Gemini 3.1 Pro & +0.54 & +0.78 & -0.24 \\
Gemini 3.7 Flash & +1.11 & +1.23 & -0.12 \\
Claude Opus 5 & +0.88 & +0.98 & -0.11 \\
GPT-5.6 Sol & +2.90 & +2.98 & -0.09 \\
GPT-5.6 Terra & +2.17 & +2.26 & -0.09 \\
Claude Sonnet 5 & +0.64 & +0.71 & -0.08 \\
Qwen 3.8 27B & +0.16 & +0.22 & -0.06 \\
Qwen 3.5 122B & +0.78 & +0.76 & +0.02 \\
\midrule
Pooled within question & +1.15 & +1.24 & -0.09 \\
\bottomrule
\end{tabular}
\end{table}
\FloatBarrier

The pooled difference between the revealed and blinded conditions is
$-0.094$ positions (95\% CI $[-0.148,-0.040]$, nominal $p=0.00068$).
Showing the generator names therefore makes judges slightly \textit{less}
likely to favor their own model, rather than more. Neither Qwen judge shows
a clear condition change: Qwen 3.5 122B has a change of $+0.022$ positions
(95\% CI $[-0.109,+0.153]$), and Qwen 3.8 27B has a change of $-0.058$
positions (95\% CI $[-0.190,+0.075]$). After Holm correction across the
eight judge-specific comparisons, only the reduction for Claude Opus remains
below 0.05; Gemini Pro has an adjusted $p=0.063$, and the other six changes
are also not significant. The overall change is small compared with the size
of the blinded effect. A shift of $-0.094$ relative to the blinded effect of
$+1.240$ accounts for roughly eight percent of the bias, and every effect in
the identity-revealed condition remains positive with an interval that
excludes zero. This suggests that self-preference is not mainly driven by
seeing the model label. Most of the effect is still present when identities
are hidden, which is more consistent with judges responding to features such
as a model's style, formatting, or reasoning conventions rather than explicit self-identification.

\subsubsection{Clinical Specialty Analysis}

If self-preference were dependent on the medical domain, it should appear in much greater magnitude in some specialties and not others. However, we found that not to be the case. All specialties had a positive ARB value, ranging from $+0.91$ positions in Radiology to $+1.54$ in Family Medicine (a difference of about $0.6$ positions). By comparison,
the differences across judges are more than four times larger, with a range
from $+0.32$ to $+2.94$. This means that the size of the effect depends more on which model is judging than on the content of what is being judged.

Specialty accounts for 5.7\% of the variation in self-preference across questions. To see whether this amount was greater than what we might expect by chance, we shuffled the specialty labels while keeping the group sizes the same. The observed value was not unusual based on this comparison ($p=0.19$), and no judge specific specialty test had a $p$-value below 0.05. Overall, this sample shows no clear evidence that self-preference depends on clinical specialty.

\subsection{Multi-Turn Results}
Similarly to the single-turn setting, models displayed clear self preference across turns. Generally, longer responses had more favorable ratings but the number of turns itself did not exhibit clear changes in the self-preference effect.

\begin{figure}[htbp]
    \centering
    \includegraphics[width=0.9\textwidth]{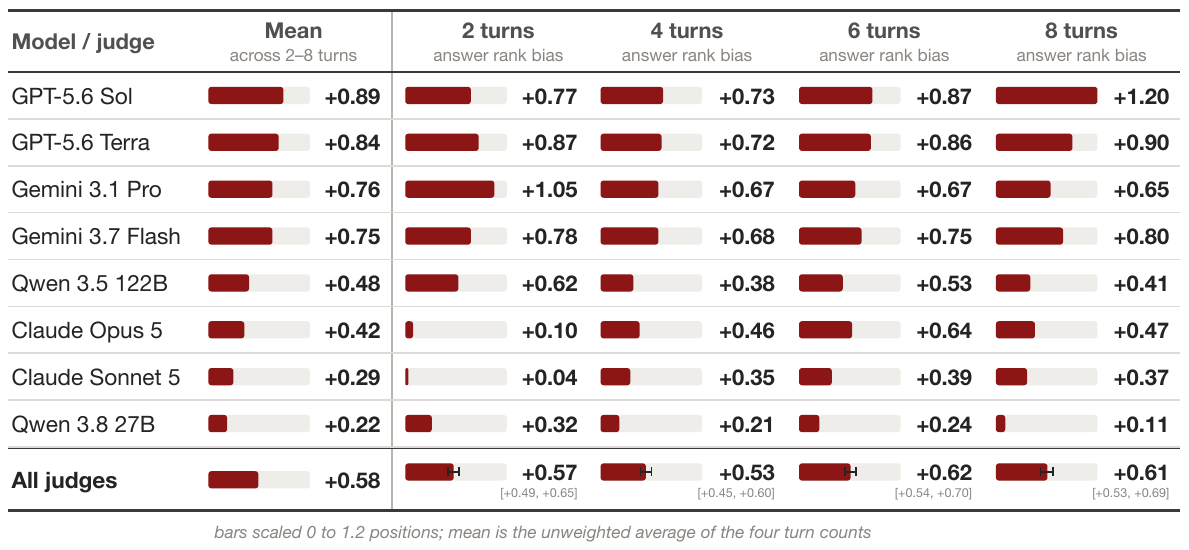}
    \caption{Single-Turn Self-Preference Results}
    \label{fig:multi-turn}
\end{figure}

There were a total of 1,600 judgments at each of the four conversation lengths with the same eight models that acted as generators and judges. At every length, the panel as a whole
ranked its own conversations higher than the outside judges ranked those same
conversations. The result also held after putting each judge's rubric scores
on a common scale (last column of Table~\ref{tab:med-pooled}).

\begin{table}[htbp]
\centering
\small
\caption{Average MedSP1000 self-preference by the number of visible turns.}
\label{tab:med-pooled}
\smallskip
\begin{tabular}{rrrr}
\toprule
Visible turns & Answer Rank Bias & 95\% CI & Standardized rubric effect \\
\midrule
2 & $+0.57$ & $[0.49, 0.65]$ & $+0.25$ \\
4 & $+0.53$ & $[0.45, 0.60]$ & $+0.25$ \\
6 & $+0.62$ & $[0.54, 0.70]$ & $+0.30$ \\
8 & $+0.61$ & $[0.53, 0.69]$ & $+0.30$ \\
\bottomrule
\end{tabular}
\end{table}
\FloatBarrier

The pooled estimate was smallest at four turns and largest at six turns, but
the differences across lengths were not clear in the overall repeated-measures
test ($F(3,597)=1.69$, $p=0.17$). A paired multivariate test led to the same
conclusion ($p=0.13$). Conversation length therefore did not show a consistent
relationship with self-preference in the eight-model evaluation.


All 32 model-by-length estimates were positive. After Holm correction, all
model-specific tests were below 0.05 except Opus and Sonnet at two turns and
Qwen 3.8 27B at eight turns. Models followed different patterns across length,
which helps explain the absence of a consistent pooled trend. The secondary measures again tell different stories:

\begin{table}[htbp]
\centering
\small
\caption{MedSP1000 secondary measures by visible length.}
\label{tab:med-secondary}

\textit{Answer Score Bias (0--5 mean-rubric scale; positive values indicate
more generous own-model scoring).}
\par\smallskip

\begin{tabular}{lrrrr}
\toprule
Model/judge & 2 turns & 4 turns & 6 turns & 8 turns \\
\midrule
Claude Opus 5   & $-0.53$ & $-0.26$ & $-0.16$ & $-0.14$ \\
Claude Sonnet 5 & $-0.48$ & $-0.14$ & $-0.10$ & $-0.10$ \\
Gemini 3.1 Pro  & $\mathbf{+0.71}$ & $+0.49$ & $+0.39$ & $+0.25$ \\
Gemini 3.7 Flash & $+0.68$ & $\mathbf{+0.59}$ & $\mathbf{+0.58}$ & $\mathbf{+0.63}$ \\
GPT-5.6 Sol     & $+0.12$ & $+0.07$ & $+0.12$ & $+0.21$ \\
GPT-5.6 Terra   & $-0.36$ & $-0.20$ & $-0.09$ & $+0.08$ \\
Qwen 3.5 122B   & $+0.52$ & $+0.31$ & $+0.30$ & $+0.28$ \\
Qwen 3.8 27B    & $+0.20$ & $+0.10$ & $+0.12$ & $+0.14$ \\
\bottomrule
\end{tabular}

\par\medskip

\textit{Own-Answer Win Rate (percentage of competing conversations ranked
below the judge's own answer).}
\par\smallskip

\begin{tabular}{lrrrr}
\toprule
Model/judge & 2 turns & 4 turns & 6 turns & 8 turns \\
\midrule
Claude Opus 5    & 51.6\% & 60.9\% & 72.9\% & \textbf{88.4\%} \\
Claude Sonnet 5  & 52.0\% & \textbf{78.1\%} & \textbf{84.6\%} & 82.5\% \\
Gemini 3.1 Pro   & 63.0\% & 50.4\% & 46.1\% & 42.8\% \\
Gemini 3.7 Flash & 52.9\% & 51.1\% & 52.6\% & 54.6\% \\
GPT-5.6 Sol      & 68.4\% & 66.9\% & 68.7\% & 77.7\% \\
GPT-5.6 Terra    & 59.8\% & 56.1\% & 58.4\% & 58.1\% \\
Qwen 3.5 122B    & \textbf{72.6\%} & 53.9\% & 43.9\% & 29.9\% \\
Qwen 3.8 27B     & 37.4\% & 34.8\% & 34.4\% & 27.5\% \\
\bottomrule
\end{tabular}
\end{table}
\FloatBarrier

The Qwen results show why these measures should remain separate. Qwen 3.5 122B
ranked first overall at two turns but fell to eighth at eight turns. Its
Own-Answer Win Rate fell with this change in conversation quality, from 72.6\% to
29.9\%. Nevertheless, its judge continued to rank its own conversation higher
than outside judges ranked that same conversation at every length. Qwen 3.8
27B also had positive Answer Rank Bias estimates at every length, although its
eight-turn interval included zero (95\% CI $[-0.06, +0.28]$). Conversely,
Opus's Own-Answer Win Rate rose from 51.6\% to 88.4\% as its conversations improved,
even though it scored its own work more harshly than outside judges did. The
Own-Answer Win Rate follows both conversation quality and judging behavior;
Answer Rank Bias holds conversation quality fixed.

\subsection{Token-length sensitivity}

Longer responses generally received better evaluations, even after comparing
answers within the same question and judge and accounting for the generator
and answer position (Table~\ref{tab:length}). In Real-POCQi, doubling length
was associated with a 1.31-place improvement in rank and a 0.59 increase in rubric score (Table~\ref{tab:length}). MedSP1000 showed the same directions at
every conversation length. 

\begin{table}[htbp]
\centering
\scriptsize
\caption{Changes associated with doubling response length. Negative
rank changes are better. The final column asks whether a model shows
more or less self-preference on its longer-than-usual answers; positive
values mean more. Brackets show 95\% confidence intervals grouped by
question.}
\label{tab:length}
\begin{tabular}{lrrr}
\toprule
Condition & Rank change & Rubric score change & ARB change \\
\midrule
Real-POCQi
& $-1.31\;[-1.43,-1.20]$
& $+0.59\;[+0.53,+0.64]$
& $-0.54\;[-0.63,-0.45]$ \\
MedSP1000, 2 turns
& $-1.66\;[-1.93,-1.39]$
& $+0.68\;[+0.56,+0.80]$
& $+0.30\;[+0.17,+0.43]$ \\
MedSP1000, 4 turns
& $-2.83\;[-3.13,-2.53]$
& $+1.15\;[+1.01,+1.29]$
& $+0.33\;[+0.19,+0.47]$ \\
MedSP1000, 6 turns
& $-3.00\;[-3.30,-2.69]$
& $+1.23\;[+1.09,+1.36]$
& $+0.29\;[+0.16,+0.42]$ \\
MedSP1000, 8 turns
& $-2.70\;[-3.04,-2.36]$
& $+1.15\;[+0.99,+1.30]$
& $+0.24\;[+0.07,+0.41]$ \\
\bottomrule
\end{tabular}
\end{table}
\FloatBarrier

The final column differs across the two tasks. In Real-POCQi, models showed less self-preference on answers that are longer than usual. In MedSP1000, they showed more self-preference on conversations that are longer than usual. Response length affects evaluation.

\subsection{Clinical validation}

Blinded physician review remains incomplete. The rankings reported here
describe how automated judges compare the answers, not how closely those
rankings agree with clinical experts.

\section{Discussion}
As medical benchmarking and monitoring relies more on LLMs-as-a-judge, it has become increasingly important to understand how LLMs exhibit self-preference. \cite{alvarezarenas2026unreliable} found that clinical LLM judges may favor AI-generated responses over human-written responses, partly because of response length and writing style.
\cite{panickssery2024llm} showed that LLM judges can recognize and favor their own generations on general tasks like news summarization, while \cite{wataoka2025selfpreference} found that judges tend to prefer lower-perplexity responses that feel more familiar to them. Our study focuses on healthcare, where benchmarking results can impact which models are selected for implementation. We examine self-preference across several model families by separating scoring generosity and direct preference for a model’s own response from Answer Rank Bias, which holds the exact answer fixed and changes only the judge. In the single turn setting, we test whether the self-preference effect remains when the rubric is removed and when generator identities are shown to the models. We also examine whether self-preference changes as multi-turn patient–physician conversations become longer, from two to eight turns.

\subsection{Single Turn using Real-POCQi}

Our experiments show that self-preference can appear differently depending on how it is measured. Every Real-POCQi judge ranks its own answer more favorably than outside judges rank that same answer. This includes models whose answers rarely rank first. For example, with the Qwen model, their answers rank first in fewer than one percent of judgments; however Qwen-as-a-judge still ranks its own answer higher than the rest of the panel does. 

The metric used to assess self-preference matters. Own-Answer Win Rate looks at pairwise comparisons and does not control for one response being objectively better (the judge picks its own answer because it is actually better). This is why we developed the Answer Rank Bias metric, which holds the answer fixed and looks at a judge's self-preference relative to a panel of outside judges. Self-preference persists after accounting for presentation position and scoring generosity. It also persists for seven of the eight judges when the rubric for ranking is removed entirely. Since judges do not know the identity of the generating model in the blind condition, the effect is better described as an affinity for a model's own output rather than recognizing their own model. This affinity may come from shared style or preferred kinds of evidence.

In our study removing the rubric did not weaken the self-preference effect. The effect is not due to judges rewarding answers that happen to match a given criteria. Together with prior work showing that self-preference survives fully objective rubrics \citep{pombal2026selfpreference}, this suggests that rubric design is not an effective tool for controlling self-preference.

When the generator names are shown, the judges still favor their own answer relative to the outside panel. In the identity revealed scenario, we would expect self-preference to increase. However, we observed that there was a slight reduction in self-preference effect. The reduction results in about eight percent of the bias, which means that hiding model identity (an obvious safeguard against self-preference) addresses very little of the effect. Therefore, explicit labels are not necessary for same-model affinity to occur and providing them does not increase it.

The family-level analysis points in the same direction. Judges ranked both answers from their family more favorably than outside judges did, by about the same amount as in the exact-model analysis. This is consistent with shared family-level tendencies. However, the calculation includes the exact-model judge, so it cannot tell us whether the other model in the family independently shows the same preference.

\subsection{Multi Turn using MedSP1000}

In MedSP1000, pooled self-preference appeared at every conversation length tested, with higher self-preference in the longer conversations (6–8 turns) than in the shorter ones (2–4 turns). However, the overall analysis did not find these differences across conversation length to be statistically significant. The largest pooled point estimate occurred at six turns, followed closely by eight turns. Individual judges follow different paths. Most notably, Qwen 3.5 122B continues to hold self-preference even as its overall quality rank falls from first at two turns to eighth at eight turns. Answer Rank Bias separates this judging preference from changes in conversation quality. These findings suggest that multi-turn conversations do not always lead to greater self-preference, and that the relationship may be more nuanced.

\subsection{Other Factors Related to Self-Preference}

Response length is a separate concern. Longer answers generally receive better evaluations, whether because of added detail, unnecessary verbosity, or both. However, the main comparison holds the exact answer fixed. In addition, models showed less self-preference on their longer Real-POCQi answers compared to their longer MedSP1000 conversations. Length affects evaluation, but it does not provide a unifying explanation for the pattern in both tasks.

We found no clear relationship between self-preference and the clinical specialty of the question. Across specialties with enough questions, the effect varies by about 0.6 rank positions. However, across judges it varies by more than 2.5 rank positions (from $+0.32$ to $+2.94$). What matters is which model is judging, not the clinical content of the question.

This is important because the two largest self-preference effects we measured belong to the GPT judges, Sol (2.94) and Terra (2.22) (Table~\ref{tab:real-self}), and OpenAI models are currently the most widely used evaluators in medical AI benchmarking \citep{li2026llm}. A benchmark that relies on a single evaluator from that family would favor answers from it, and researchers may not realize the role self-preference is playing. Medical benchmarking studies should therefore report results by judge and use multiple judge families rather than relying on a single pooled evaluator. Since new frontier models are released frequently, we provide a continuously updated leaderboard https://daneshjoulab.github.io/Medical-LLM-Self-Preference/ with the latest results.

\section{Limitations}

There are a few notable limitations to the work. Each model generated one response per question or scenario, and each judging condition was run once. The study therefore does not quantify variation across repeated generations or judging runs. We also did not benchmark the judges against physician evaluations, so in some cases a judge may prefer its own answer because that answer is clinically favorable. We attempt to control for this by comparing rankings/evaluations across a range of models, but follow up human-expert evaluation would strengthen the analysis. We also attempt to deepen understanding of self-preference effects in relation to token length, same model-family affinity, and different grading procedures. However, each of these areas was done on subsamples of our initial dataset and would benefit from further analysis.


\section{Conclusion}

Our findings show that self-preference is present in medical LLM judging, but it does not appear in the same way across all models or measures. The experiments show that LLM judges rank answers from their own exact model more favorably than other judges rank those same answers. This pattern remains across two medical settings, after adjusting for presentation position, using rubric scores, and in a no-rubric sensitivity analysis. In multi-turn conversations, pooled self-preference was present at every tested length, without a clear overall difference across lengths. Revealing generator names slightly reduces Answer Rank Bias while still leaving strong same-model affinity. These findings highlight the importance of using multiple judge families, multiple evaluation measures, and fixed-answer analyses when comparing medical LLMs.

\bibliographystyle{unsrtnat}
\bibliography{references}

\end{document}